\documentclass[letter, 10pt, conference]{ieeeconf}      % Use this line for a4 paper

\IEEEoverridecommandlockouts                              % This command is only needed if 
\usepackage{graphics} % for pdf, bitmapped graphics files
\usepackage{epsfig} % for postscript graphics files
\usepackage{mathptmx} % assumes new font selection scheme installed
\usepackage{times} % assumes new font selection scheme installed
\usepackage{amsmath} % assumes amsmath package installed
\usepackage{amssymb}  % assumes amsmath package installed
\usepackage{color}
\usepackage{balance}
\usepackage{siunitx}
\usepackage{amsmath}
\newcommand\inlineeqno{\stepcounter{equation}\ (\theequation)}

\title{\LARGE \bf
Domain Independent Unsupervised Learning to grasp the Novel Objects
}

\author{Siddhartha Vibhu Pharswan$^{1,2}$, Mohit Vohra$^{1}$, Ashish Kumar$^{1}$ and Laxmidhar Behera$^{1}$ % <-this % stops a space
\thanks{$^{1}$Intelligent Systems and Control Lab (ISL) Indian Institute of Technology Kanpur, Uttar Pradesh 208016, India. {\tt\small vibhu@iitk.ac.in, mvohra@iitk.ac.in, krashish@iitk.ac.in, lbehera@iitk.ac.in }}%
\thanks{$^{2}$Department of Mechanical Engineering  Indian Institute of Technology Kanpur, Uttar Pradesh 208016, India.}%
}

\begin{document}
\maketitle
\thispagestyle{empty}
\pagestyle{empty}

\begin{abstract}
One of the main challenges in the vision-based grasping is the selection of feasible grasp regions while interacting with novel objects. Recent approaches exploit the power of the convolutional neural network (CNN) to achieve accurate grasping at the cost of high computational power and time. In this paper, we present a novel unsupervised learning based algorithm for the selection of feasible grasp regions. Unsupervised learning infers the pattern in data-set without any external labels. We apply k-means clustering on the image plane to identify the grasp regions, followed by an axis assignment method. We define a novel concept of Grasp Decide Index (GDI) to select the best grasp pose in image plane. We have conducted several experiments in clutter or isolated environment on standard objects of Amazon Robotics Challenge 2017 and Amazon Picking Challenge 2016.  We compare the results with prior learning based approaches to validate the robustness and adaptive nature of our algorithm for a variety of novel objects in different domains.

\end{abstract}

\section{INTRODUCTION}

A robot with grasping capability has tremendous applications in warehouse industries, construction industries, or in medical sector. Various solutions have been proposed to enhance the grasping capability of manipulators. Authors \cite{pas15}, proposed geometric methods to predict the grasp points for unknown objects present in a clutter. In \cite{lev17} \& \cite{pin16}, authors collected large scale data set to train a giant convolutional neural network (CNN) to predict the grasp regions. Although CNN-based solutions have shown some good results, but their performance depends on the data set. Thus making these solutions data biased and domain (background) dependent. The effect of domain specific nature can easily be examined by taking, an example of a network trained on one particular background \cite{lev17}, but can take equal or less training time when transferred to the different background. Selecting proper hyper-parameters can reduce the training time but still the difference is not that significant. Consequently, robot grasping is dealing with generalization at the cost of high computation power and time.
\begin{figure}[!h]
	\centering
	\includegraphics[width=8.15cm,height=12cm]{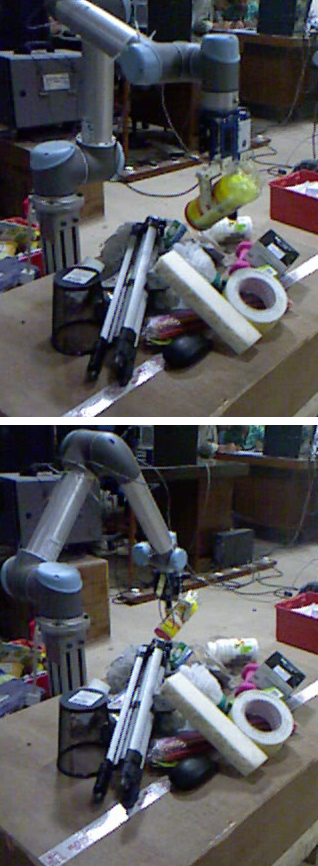}
	\caption{Our robotic hardware grasping object in dense clutter. The robot is grasping the object after the final selection is done by GDI in one of the experiments.}
	\label{fig1}
\end{figure}
\setlength{\textfloatsep}{7pt}
When dealing with novel objects unsupervised nature like human beings is needed in algorithm to identify the grasp regions. This behaviour involves the single shot screening of work-space with all the grasp regions along with final grasp. While the current approaches rely on creating or accumulating data, our method is not data specific. We structured the algorithm using an unsupervised learning baseline to find out the grasp regions in a clutter. Collision phenomenon \cite{ulr17} is used to avoid the gripper collision with surrounding objects. We represent the two-finger gripper as a rectangle \cite{jia11} in the image plane as shown in Fig. 2. In this representation, the opening of the gripper is represented by the length of the red line. Green line and centroid point represent the two fingers and palm of the gripper. The advantage of representing gripper in the image plane allows us to limit the search space also, it helps in identifying whether a grasp pose is feasible or not. Using the point cloud data corresponding to the sampled points inside gripper rectangles, collision with other objects is avoided or completely eliminated. For finding the feasible grasp regions, we divide our algorithm into 5 stages where each stage is hierarchically related to its previous state. In the first stage, poses are sampled uniformly at random orientations in the image plane. After sampling, poses which belong to no object regions are removed. These stages are followed by cluster and axes assignment and finally, the most feasible grasps are selected by GDI.

One of the significant contributions of our work is, neither our work requires any pre-processing step, nor does it need any prior information about the objects. We are thus providing a generalized solution. Besides this, our framework is light enough to run on a single computer making it an economical solution.

\section{RELATED WORK}
To solve the robot grasping problem, numerous solutions have been proposed for the last decades. Most of them are focused on known object grasping, while only a few are more concerned about the novel object grasping \cite{pin16}, \cite{lev17}. In \cite{sax08}, authors uses machine learning approaches to predict the grasp points on two images of the object, and the points are triangulated to predict the 3D location of the grasp point. The results in \cite{sax08} are quite specific and not even shown in real practical clutter. In \cite{katz14}, authors estimate the surface of unknown objects in a clutter using depth discontinuities in depth image and normal vector at each point, and train a network to predict the grasp regions for unknown objects in a pile.\cite{bou15} uses Voxel Cloud Connectivity Segmentation method to detect surface of the object and proposes a reinforcement learning based system which can learn to manipulate the objects by trial and error. \cite{chit12} have done modelling of robot grasping environment to perform the manipulation task. While this technique completely relies on the combination of multiple sensors with in the grasping system. Key advantage of our method is, neither we require any segmentation techniques nor we need any normal vector for surface estimation, thus making our approach less complex. 

Because of high learning capability and generalize nature of CNN, various CNN-based methods \cite{red15}, \cite{sul17} have been proposed which predict the grasp region directly using RGB images. Key requirement of CNN-based methods is the large scale data-set. In \cite{lev17},  to show the generalization of the grasp framework, authors use 800000 grasp attempts and then transfer the same strategy to other robots. \cite{pin16} shows the generalization of the framework by 700 hours of robot training. Authors \cite{her14} have proposed a template based learning approach, which depends upon the creation of template data for set of objects and make it generalise on other similar or dissimilar objects. While this approach also affects the same problems that CNN faces like lightening conditions, data creation strategies etc.

In \cite{mah16}, authors use a large data-set of 10000 3D models, with 2.5 million gripper poses for training. The grasp metrics and data-set for grasping objects is proposed by \cite{boh15}. In \cite{len15}, \cite{red15}, authors uses CNN-based architectures to predict the rectangular grasp regions in the image plane. In \cite{keh13}, authors present a system architecture for a cloud-based robot grasping which uses massive parallel computation power and real-time sharing of vast data resource. They used google object recognition system followed by creating 3D models in offline mode for analyzing the object poses and grasp regions. During testing, image of object is sent to object recognition server for pose estimation and grasp region selection. A more related review of previous work in the field of grasping is given in \cite{boh16}.

While all these works are creating a bunch of data to solve the problem, our approach focuses on finding the top feasible grasp poses.
\begin{figure}[!h]
	\centering
	\includegraphics[width=6.75cm, height=6cm]{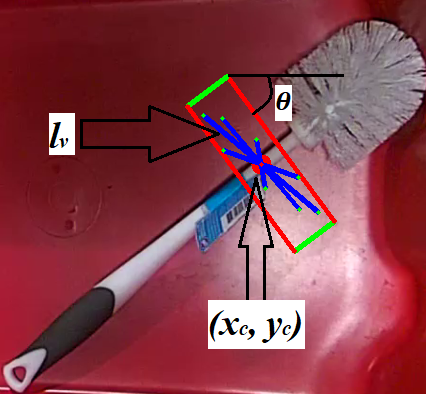}
	\caption{Gripper representation as rectangle and lines of varying lengths on the object to be grasped. The length of red line (gripper opening) is maximum opening of gripper mapped to image plane. Green points on every line gripper are the gripper fingers in image plane.}
	\label{fig2}
\end{figure}
\setlength{\textfloatsep}{8pt}
We have shown in the experiments that just by single Kinect sensor how grasping can be solved robustly at less cost. The way we have posed the problem, I think is the first time unsupervised learning in the whole pipeline is used without any labelled data-set and training.  

\section{PROBLEM STATEMENT}
Given an image \textbf{\textit{I}} and corresponding point cloud \textbf{\textit{P}} $\mathbf{\subset \ R^3}$  of the robot task-space from any RGB-D sensor (in our case Kinect V1 of resolution 640 $\times$ 480), how to decide the feasible grasp regions without any prior knowledge of objects. The main step in our problem is to find the object and no-object regions in the image plane. We used a sampling based strategy to find the object and no-object regions which exempt us to use any standard segmentation techniques.
\par
% We have conducted all the experiments on Universal Robot 5 (UR5) having two finger gripper with Kinect V1 attached at the top of the robot workspace. Gripper has maximum opening angle of $180^o$. Robot Operating System (ROS) is used to integrate the whole pipeline and Rapidly-expanding Random Trees (RRT) is used for task space planning to desired point.

\subsection{Gripper Representation in Image Plane} 
Depending on the stage of our algorithm, we represent the gripper either by a rectangle or by a line. For sampling stage, gripper is represented by a line having parameters ($x_c,\ y_c,\ l_v,\ \theta$) and for GDI calculation, it is treated as a rectangle with three parameters ($x_c,\ y_c,\ \theta$). Here $x_c$, $y_c$ are the palm co-ordinates of gripper in image plane, $l_v$ is the length of line segment which represents the opening of gripper in image plane, $\theta$ is angle of rectangle or line with horizontal axis. Fig. \ref{fig2} shows gripper representation. To decide the optimum opening of gripper in image plane for known objects, an object of maximum width is selected for mapping along with additional clearance. On the other side, to pick up the novel objects gripper opening is decided on the basis of task category. For small household objects, opening of gripper is kept half of its' maximum opening. Width of gripper also depends on the geometric constraints of gripper design as well. We used the gripper with maximum opening of \ang{180}. To avoid collision of gripper fingers with near by objects, clearance of 2cm (mapped to image plane) is provided at each side of gripper finger. 

\subsection{Sampling and Filtering}
To grasp an object, either robot has prior object information [16], [17] or it has learnt by trials [6], [7]. In real time situations, we do not have any prior information of the objects. To identify the object and no-object regions, previous methods \cite{pin16}, \cite{lev17} either trained models rigorously or prefer object modelling \cite{qui09}. In our method, we used the depth information of the work-space from the 3D sensor (Kinect in our case). Given a work-space, we perform uniform sampling of the line configurations of the gripper in the image plane within the task-space of robot. These configurations are at random angles with fixed $l_v$ (depends on maximum width of object) for a particular clutter. This sampling stage is the $1^{st}$ stage of the algorithm. Large number of line configurations are sampled to cover the entire task-space in image plane.
\par
As all the sampled configurations will not be over object regions, to remove the false samples we perform filtering operation in two levels. In first level, samples belong to no-object regions are removed. Here, we filtered the false poses using the Z-values of center point of line configurations and task-space surface. As the center point of every line pose represents the palm of the gripper, so to grasp an object firmly, the z-value of center point must be greater than the z-value of background (domain where objects are placed). So we will select those poses whose Z-value of centre points are greater than the background z-value. In the second level, we remove the samples which have high probability of object gripper collision. In the gripper representation, corner points of line represent the finger of the gripper. If difference between the z-values of corner points is large, it means one of the finger is in collision with the objects, hence we will reject these poses as well. Above conditions can remove all the false poses when objects are placed separately, but in a clutter environment still there will be some false poses left which will be removed in further stages. Output of filtering stage helps in segmenting the object and no-object regions using point cloud information only, which makes the algorithm to work in any lightening conditions as well. This nature of domain independence is inherited from the nature of 3D sensor used.
\subsection{K-Means Clustering}
Once the configurations in object regions are obtained, next step is to group the particular set of configurations and localise it in image plane. For this task, we apply k-means clustering on the remaining configurations. The cluster centroids are taken from the center points of gripper configurations. Center points are taken as input because it forcibly allows the k-centers to lie on the objects surface.  If corner points were taken as input to the k-means clustering algorithm, then the sampled poses present at the corner sides of objects would be grouped together by k-means clustering. Thus the chance of local minima is more. Local minima can be seen in the following common situations \emph{i)} sampled poses on two objects are identified by a same k-center \emph{ii)} more than two k-centers on the same pose family. Later situation is more effective because in robot grasping, that case helps to grasp the object in multiple ways. So selecting centre points as input can minimize the chance of local minima but it can not be avoided in case of dense clutter. In experiment section we demonstrate that performance of our algorithm is not effected by local minima due to the ranked selection of grasps.
\begin{figure}
	\includegraphics[width=8cm,height=6cm]{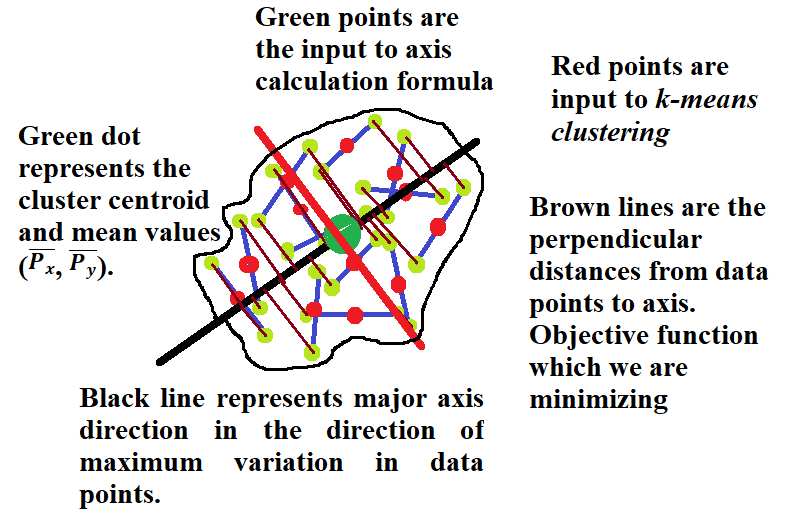}
	\caption{Axis assignment for one of the point families cluster. Contour line represents the region of poses belongs to the index of green cluster centroid.}
	\label{fig3}
\end{figure}
\subsection{Axis Assignment}
K-means clustering divides the object region into multiple segments, and every segment has a set of line posses. To assign axes to a single segment, the corner points of line poses which are present inside that segment, will be used. For each segment, the corner points of all the line poses are treated as points family. 

So if $k^{th}$ segment has S poses, then segment will have 2S (T) corner points and $k^{th}$ cluster will have a point family of 2S points. For each point family, axis assignment (major axis angle $\phi$) is done using the formula[5] given below. Let $\mu_x,\ \mu_y$ represents the centroid of cluster having T corner points. Let us assume that $i^{th}$ corner point can be represented as ${P^i}_{kx},\ {P^i}_{ky}$ and $\phi$ is the angle of major axis with horizontal axis. Relation between centroid, corner points and major axis angle is given by 
% The axis assignment relies on the hypothesis to grasp the object along the major axis just to fit it inside the gripper [13].
$$tan(2\phi) = 2* (\frac{\sum_{i=1}^{T} ({P^i}_{kx} - \mu_x)*({P^i}_{ky} - \mu_y)}{\sum_{i=1}^{T} [({P^i}_{kx} - \mu_x)^2 + ({P^i}_{ky} - \mu_y)^2]})   \inlineeqno $$ When objects are placed separately, major axis will lie along the major axis of the objects. Axis assignment is described on Fig. \ref{fig3} using one point family and its associated cluster centroid.

\begin{figure}
   \includegraphics[width=8.5cm,height=4cm]{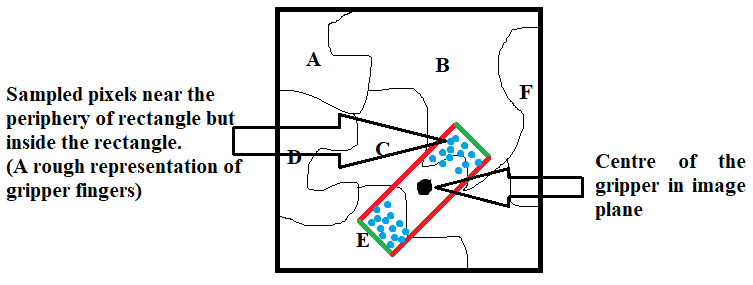}
   \caption{Sampled pixels inside the rectangle representing finger of gripper in image plane over point regions. A,B,C etc are the point family clusters for corresponding cluster centroid (black dot) in image plane.}
   \label{fig4}
\end{figure}\setlength{\textfloatsep}{12pt}

\begin{figure}
   \includegraphics[width=8.75cm,height=4cm]{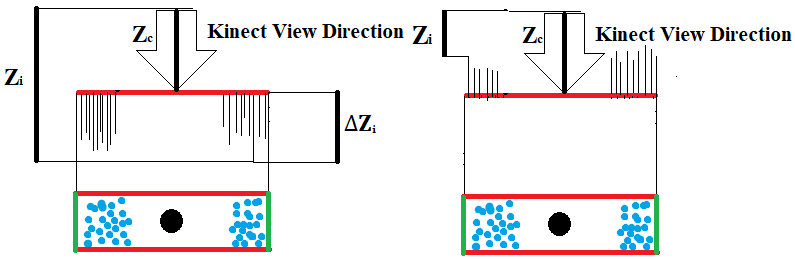}
   \caption{Top and side views of extreme GDI cases in practical conditions. Top view shows the sampled points inside rectangle in image plane and side view represents z-values corresponding to them. Intermediate cases gets cancelled from the above formula.}
\label{fig5}
\end{figure}

\subsection{Grasp Decide Index}
In axes assignment stage, we assign a major axis to each segment. But output of above stage leads to k-axes groups assigned to every point family. It is because the direction of major axis is decided by the point family of a segment, and in a clutter environment it could be possible that more than one object could be the part of that segment. To filter out the false assignments and select the final graspable assignment, we propose an index. This index takes into account the collision of gripper fingers with surrounding objects. For GDI calculation, we represent every cluster by a rectangle with centre at centroid of cluster, and orientation of rectangle is $\theta$ as $\phi + 90$. To represent the fingers of gripper in image plane, we have sampled points near rectangle periphery but inside it as shown in Fig. 4. Let the z-values of $i^{th}$ sampled pixel and rectangle center (palm center) is $Z_i$ and $Z_c$ respectively. The number of positive deviations ($\Delta Z = Z_i - Z_c$) are different for every rectangle in image plane. The more the number of positive deviations, a rectangle has less the chance of collision. The Grasp Decide Index (GDI) is formulated as follows:
$$ GDI = \max_{N} (Z-Z_c),\;where\;Z \in R\,^N   \inlineeqno $$   
\emph{\textbf{N}} is the number of sampled pixels in each rectangle, \textbf{\emph{max}} denotes the maximum positive deviation for a rectangle over N-pixels. Final rectangle out of those is selected on the basis of one with highest GDI. Avoiding the random grasp selection strategy, index ensures the safety of grasp which can be seen in Fig. 5.
\section{EXPERIMENTS AND RESULTS}
\subsection{Objects Used for Grasping}
We have used total 50 objects out of which 35 were standard objects from Amazon Robotics Challenge 2017 and Amazon Picking Challenge 2016. Remaining objects are household and office stationary objects. These objects are placed randomly in clutter or separately on different domains for all the experiments. We have used two different backgrounds i.e. red bin and wooden table. Fig. 6 shows a major part of object set used for the experiments.
\begin{figure}
	\centering
	\includegraphics[width=8cm, height=6cm]{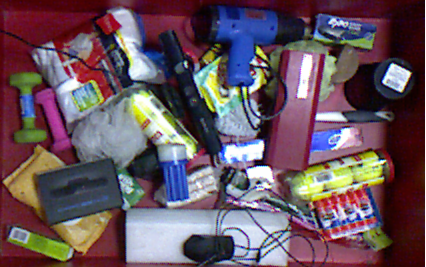}
	\label{fig6}
	\caption{30 objects out of 50 in red bin. These objects have varities of difficulty in shape and size.}
\end{figure}\setlength{\textfloatsep}{10pt}
\subsection{Parameters Selection and Positive-Negative Effect}
In second stage, to remove the false poses we perform two level of filtering using depth value. Since we placed the Kinect at a height of 1.3m above the workspace and because of noisy point cloud, we add a depth margin value of 0.025m for filtering. In our experiments we found that, this margin is sufficient to grasp the thin objects (around 2.5m thick) and also we can decrease this margin value by placing Kinect nearer to workspace. 
\begin{table}[!h]
	\centering
	\caption{Results of cluttered environment}
	\label{tablec1}
	\begin{tabular}{|c|c|c|c|c|c|}
		\hline
		Environment & NoT & OP & MT & $\alpha$($\%$) & $\beta$($\%$)\\
		\hline
		Clutter(Red) & 11 & 10 & 12 & 90.90 & 91.67\\
		\hline
		Clutter(Wood) & 20 & 18 & 22 & 90 & 90.9\\
		\hline
		Clutter(Wood) & 15 & 13 &17 &86.67&88.24\\
		\hline
		Clutter(Wood) & 15 & 14 & 16 & 93.33 & 93.75\\
		\hline
		Clutter(Red) & 10 & 8 & 13 & 80 & 76.92\\
		\hline 
	\end{tabular}
\end{table}
\setlength{\textfloatsep}{10pt}
\begin{table}[!h]
	\centering
	\caption{Results of no clutter}
	\label{tablec2}
	\begin{tabular}{|c|c|c|c|c|c|}
		\hline
		Environment & NoT & OP & MT & $\alpha$(\%) & $\beta$(\%)\\
		\hline
		Seperated(Red) & 15 & 14 & 16 & 93.33 & 93.75\\
		\hline
		Seperated(Wood) & 15 & 15 & 15 & 100 & 100\\
		\hline
		Seperated(Red) & 15 & 14 &16 &93.33&93.75\\
		\hline
	\end{tabular}
\end{table}

\begin{figure*}
	\includegraphics[width=17.5cm,height=13cm]{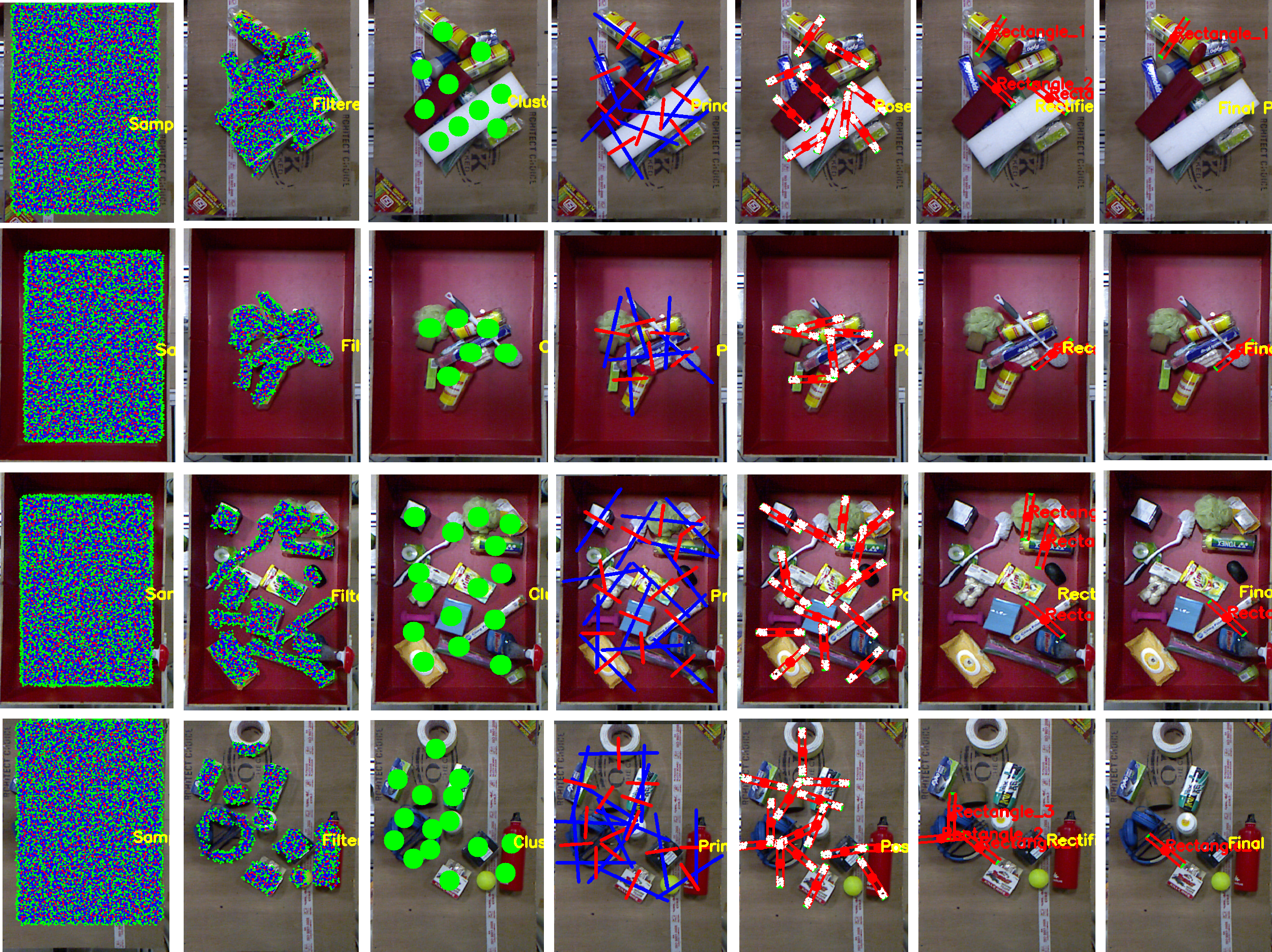}
	\label{fig7}
	\caption{All the stages of the algorithm running in image plane. Objects are placed on different backgrounds with different conditions (isolation or clutter).}
\end{figure*}

In our experiments, we found that k-means clustering methods stuck in local minima. This situation occurs when we try to assign a single centroid to more than one family points, or when we assign multiple centroids to a single point family. But the performance of our system is not affected because in the first case, the grasp pose will be removed by the GDI stage, while in second case multiple centroids to a single family provide various ways to grasp the object. In all our experiments value of K is taken as 6 to 12 without any consideration of how many objects are there in bin because as the grasping proceeds clutter will decrease and objects can be grasped easily (latter case of local minima). While for the case of isolation, K is equal to number of objects present in the task-space. All stages start from the sampling to final pose selection are shown in Fig. 7 for two different domains. We have tabulated the results of our experiments in the Table I and Table II for clutter and no clutter respectively where number of trials (NoT) robot take is equal to number of objects present.

\begin{figure*}[!h]
	\includegraphics[width=17.5cm,height=11cm]{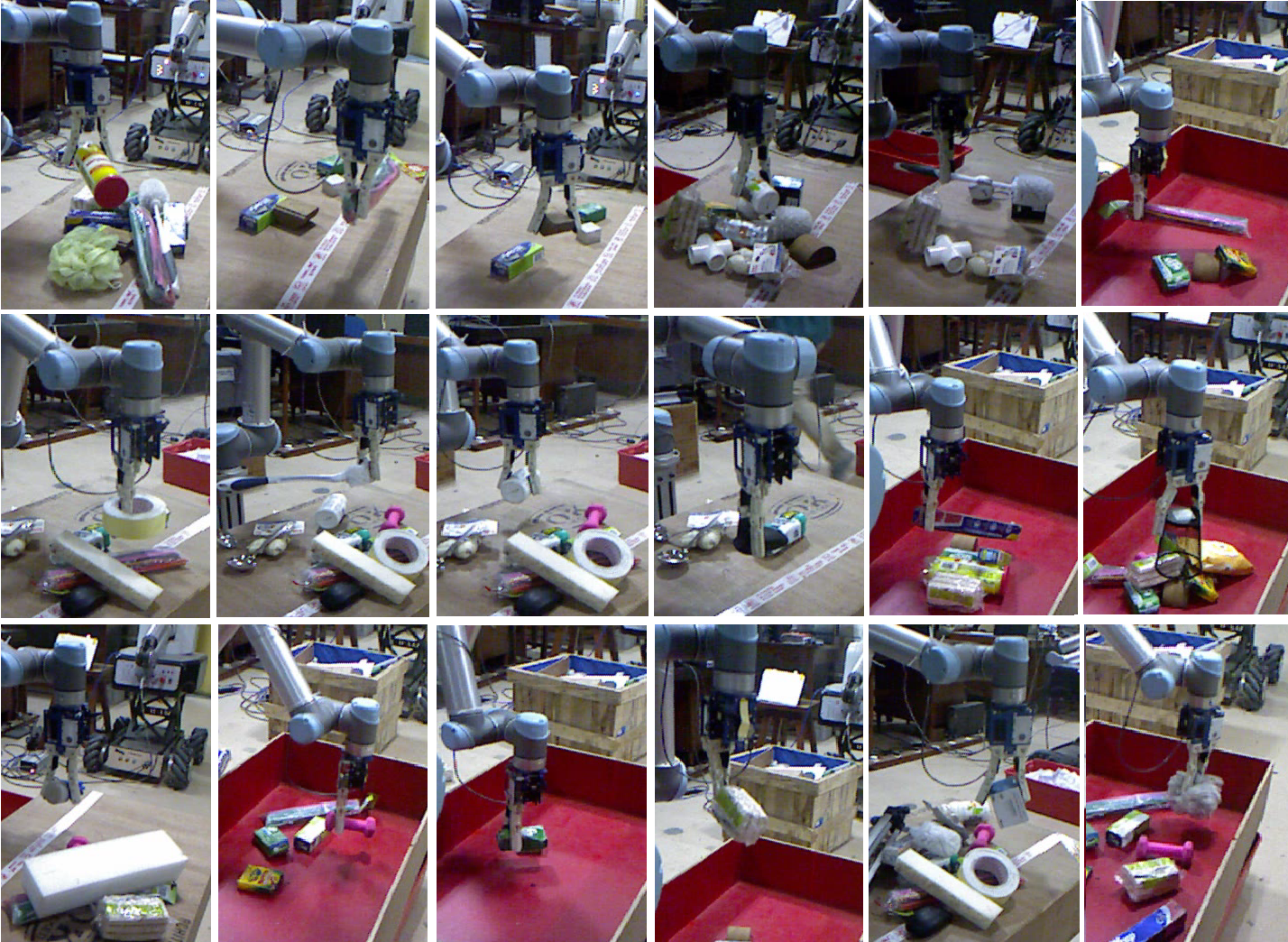}
	\label{fig8}
	\caption{Robot grasping the variety of objects in cluttered and uncluttered environment.}
\end{figure*}

Total 8 experiments are conducted. For five experiments, objects were placed in clutter inside red bin and on wooden table. Remaining three experiments are performed for uncluttered domains(red bin or wooden table). For each uncluttered experiment, objects were placed separately in random pose, even sometimes in overlapping situations, while for cluttered experiments objects are placed randomly in the bin or table. For the visualization, we select the top 5 poses with rank 1 to rank 5 based on their decreasing GDI values respectively. Final selected pose is the rank 1 pose having highest GDI among the selected poses and has minimum chance of collision.
To measure the performance of our algorithm, we have used three main factors i) maximum robot trials to clear the task-space (MT), ii) object picked (OP), iii) number of trials (NoT) which are equal to the number of objects present in the task-space. Two parameters $\alpha = OP/NoT$ and $\beta=NoT/MT$, measure the grasp success rate on the basis of robot trials. $\alpha$ strictly focuses on slipping failure and algorithm bad prediction cases. Percentages of objects picked over number of trials (NoT) and objects picked over maximum trials (MT), when average over all clutter experiments show the accuracy of 88.18\% and 88.292\% respectively. Table II shows the result of no-clutter cases where the $\alpha$ and $\beta$, averaged over all experiments are 95.55\% and 95.83\% respectively. Our accuracy measure of experiments is considered taking NoT as the reference because domain will obviously be cleared for some extra trials which is the case with $\beta$. We have not tried rigorous trials to empty the work-space like \cite{pin16}. 

\subsection{Comparisons with past grasping techniques}\vspace*{-0.1em}
We have compared our results with random action strategy, heuristics strategies like i) grasping near the centroid, ii) grasping along minimum eigen axes (which is reflected just to fit the object in gripper) and iii) grasping the top in clutter [13]. In learning based strategies, we have compared the results of our algorithm with that of \cite{pin16}. In experiments \cite{pin16}, the algorithm has high accuracy on train set but less at test set, so to make valid comparison we have taken average of accuracy of train set and test set.
\begin{table}[!h]
	\centering
	\caption{Comparative study of grasping}
	\label{table5}
	\begin{tabular}{|c|c|}
		\hline
		Random Strategy\cite{katz14} & 15\% \\
		\hline
		Heuristic Strategies\cite{katz14} & 92\% (No clutter), 40\% (Clutter)\\
		\hline
		Learning Strategy\cite{pin16} & (40K trials) 86.3\% \\
		\hline
		Proposed Strategy &	95.5\% (No Clutter), 88.18\% (Clutter)\\
		\hline
	\end{tabular}
\end{table}
Their results are published on 150 variety of objects and we had also tested our algorithm on standard and similar type of objects like brush, headphones, hand- drill etc. We have compared the $\alpha$ value of our experiments with these strategies as it reflects the real-effect of algorithm prediction nature. We have not focused on maximum number of trials percentage as it is more specific to empty the bin instead of algorithm success.

\begin{figure}[!h]
	\includegraphics[width=8.5cm,height=4.8cm]{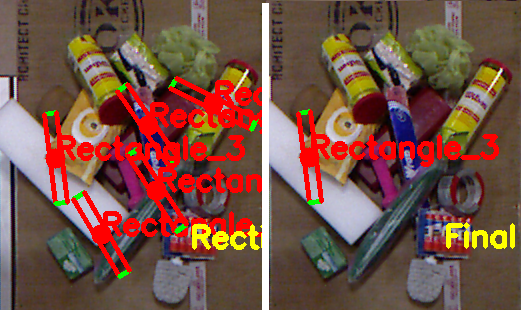}
	\label{fig9}
	\caption{In this situation the safe grasp is selected out of all the grasp but this grasp will not lead to successful attempt causes more attempt.}
\end{figure}

\subsection{Causes of success and failure}
It has been found that our method is biased on sampling. More dense sampling will lead to better results. During our initial experiments we sampled very less poses with $l_v$ taken as 80 and got very poor results in clutter. Improvement in results is achieved by reducing the value of $l_v$ to 30 (in pixel units) with dense sampling. If objects are placed separately these values does not matter too much. Another noticeable fact is, GDI includes the safety measure, it does not consider the success of grasp. GDI, mostly selects the collision free grasps without considering its' success for a particular shape of the object. In some cases, if k-center gets assigned to two clusters or vice-versa, then the axes assignment step will be the cause of failure. As, the axes assignment is done blindly, it will only consider the effect of high deviation within the point families, which may result in a some axes assignment in no object regions or along the minor axes of object itself as shown in Fig. 8.

\section{CONCLUSIONS}
We have proposed a novel real time grasp pose estimation technique using unsupervised learning baseline. To demonstrate the performance of our grasp framework, we have conducted several experiments presented in this paper. Starting from the sampled poses, our framework finds the best poses in clutter with high percentage of success. Our method does not rely on any standard segmentation techniques which allows it to deal with any background. If we consider only the removal of clutter in our experiments then our algorithm will outperform in comparison to previous methods as shown in all the experiments.

\addtolength{\textheight}{-10cm}

\end{document}